\documentclass{article}

\usepackage[preprint]{neurips_2025}

\usepackage[utf8]{inputenc}    
\usepackage[T1]{fontenc}       
\usepackage{amsmath,amssymb}   
\usepackage{graphicx}          
\usepackage{xcolor}            
\usepackage{booktabs}          
\usepackage{microtype}         
\usepackage{url}               

\usepackage{algorithm}
\usepackage{algpseudocode}
\usepackage{listings}

\usepackage{authblk}
\title{Token Constraint Decoding Improves Robustness on Question Answering for Large Language Models}

\author[1]{Jui-Ming Yao (Raymin Yao)}
\author[2]{Hao-Yuan Chen (Mark Chen)}
\author[1]{Zi-Xian Tang}
\author[1]{Bing-Jia Tan}
\author[3]{Sheng-Wei Peng}
\author[1]{Bing-Cheng Xie}
\author[3]{Shun-Feng Su}

\affil[1]{%
  Department of Computer Science and Information Engineering\\
  National Taiwan University of Science and Technology, Taipei, Taiwan%
}
\affil[2]{%
  Bachelor’s Program in Computer Science\\
  University of London, London, United Kingdom%
}
\affil[3]{%
  Department of Electrical Engineering\\
  National Taiwan University of Science and Technology, Taipei, Taiwan%
}

\affil[ ]{%
  \texttt{\{b11132009, b11115001, b11115023, b11130009, m11207330, sfsu\}@mail.ntust.edu.tw}, 
  \texttt{hc118@student.london.ac.uk}%
}

\begin{document}

\maketitle

\begin{abstract}
Large Language Models (LLMs) have demonstrated impressive performance on multiple-choice question answering (MCQA) benchmarks, yet they remain highly vulnerable to minor input perturbations. In this paper, we introduce and evaluate Token Constraint Decoding (TCD). This simple yet effective inference-time algorithm enforces alignment between token-level predictions to enhance robustness in noisy settings. Through extensive experiments on CommonsenseQA, MMLU, and MMLU-Pro, we show that TCD, especially when paired with prompt engineering (PE) fixes, significantly restores performance degraded by input noise, yielding up to +39\% absolute gains for weaker models like Gemma3 1B. Penalty sweep analyses further reveal that TCD implicitly regularizes overconfident outputs, with different models requiring distinct penalty schedules to maximize resilience. Our findings establish TCD as a practical, model-agnostic approach for improving reasoning stability under real-world imperfections and pave the way for more reliable deployment of LLMs in safety-critical or user-facing applications.
\end{abstract}

\section{Introduction}
Large language models (LLMs) such as Gemma series from Google \cite{gemmateam2024gemma2improvingopen, gemmateam2024gemmaopenmodelsbased}, LLama models from Meta AI \cite{touvron2023llama2openfoundation}, and GPT series from OpenAI \cite{brown2020languagemodelsfewshotlearners} have demonstrated remarkable capabilities in question answering and conversational tasks \cite{robinson2023leveraginglargelanguagemodels}, significantly expanding the scope of natural language processing. However, the work identifies a critical vulnerability within LLMs – their susceptibility to identifying human-readable noise, specifically spacing irregularities between options or descriptions \cite{zheng2024largelanguagemodelsrobust}. We found that appending a space after a control keyword, such as "Answer:", can substantially degrade model performance when evaluated using exact-match metrics.  This fragility represents a fundamental robustness failure in LLMs, where semantically equivalent inputs often yield divergent outputs. For instance, the alteration of a control keyword, such as "Answer: ", can lead to a noticeable drop in accuracy, even when the altered prompt maintains the same underlying intent. Our experiments across models like LLaMA \cite{touvron2023llama2openfoundation} and Gemma \cite{gemmateam2024gemma2improvingopen, gemmateam2024gemmaopenmodelsbased} demonstrated that these perturbations significantly reduce exact-match accuracy, even when the altered prompt conveys the same intended meaning. These results highlight a concerning trend: subtle spacing variations, while imperceptible to humans, have a measurable and sometimes substantial impact on LLM performance.

Therefore, the study proposes a novel decoding method that significantly resolves such robustness issues for LLMs on question answering tasks such as CommonsenseQA \cite{talmor-etal-2019-commonsenseqa} and MMLU \cite{hendrycks2021measuringmassivemultitasklanguage} benchmarks. The method, Token Constraint Decoding (TCD), involves constraining the model's output to a specific set of allowed tokens, ensuring that the generated text remains relevant and contextually appropriate. By defining a list of permissible tokens, the method modifies the model's output logits to select only these tokens. It applies penalties to disallowed tokens and uses temperature scaling to adjust the randomness of the output. Additionally, the method includes an optional debugging feature to track and analyze token changes, providing insights into the impact of these constraints. Overall, Token Constraint Decoding offers a way to achieve more controlled and accurate outputs from language models. The experiments have demonstrated a significant improvement in the robustness of LLMs on such issues, including multi-modal and common-sense reasoning benchmarks. 

\section{Related Works}
Several studies have explored the robustness of large language models (LLMs) to various types of textual perturbations \cite{khatun2024studylargelanguagemodels, li-etal-2025-prdetect}, revealing consistent vulnerabilities and differences across model types \cite{zheng2024largelanguagemodelsrobust}. Misspellings, even minor ones like changing ``library'' to ``libarary'' \cite{gan-etal-2024-reasoning}, significantly reduce model performance across both small and large LLMs. Although larger models offer some resilience, multilingual models like mT5 \cite{xue2021mt5massivelymultilingualpretrained} show superior robustness to real-world typos. Word order perturbations, such as swapping the sequence of subjects and objects, generally do not affect the semantic understanding of models like GPT, which demonstrate robustness in these cases \cite{brown2020languagemodelsfewshotlearners}. However, models such as LLaMA are more sensitive and show degraded performance. Introducing irrelevant context \cite{shi2023largelanguagemodelseasily}, like injecting unrelated sentences into a prompt, can mislead models and sharply increase error rates---especially in tasks like math problem solving. 

Yet, strategically prompting the model to ignore such noise can mitigate these effects. When it comes to semantic obfuscation, such as using complex synonyms or paraphrases, models often struggle to grasp the intended meaning, especially if the expression becomes overly obscure. Instruction tuning may worsen robustness in these cases. Finally, syntactic and punctuation errors---ranging from misplaced commas to verb form mistakes---are somewhat tolerated by advanced models like GPT-4 \cite{openai2024gpt4technicalreport}, which can self-correct minor issues. However, severe grammatical corruption still disrupts comprehension, particularly in less robust or insufficiently trained models. Overall, while larger and more advanced LLMs show improved fault tolerance, they remain susceptible to specific perturbation types, underlining the need for targeted robustness training.

\section{Token Constraint Decoding (TCD)}
Therefore, robustness remains a critical challenge for large language models (LLMs) in real-world applications. While these models demonstrate impressive capabilities across a range of tasks, their reliability often degrades in the face of minor, human-like perturbations, such as typos or formatting inconsistencies, that do not affect semantic meaning. To address this, we propose \textit{Token Constraint Decoding} (TCD), an inference-time technique designed to enhance model stability and control without requiring retraining. Firstly, the problem is mathematically formulated and analyzed in the section, and a novel decoding strategy and complementary methods to further improve the robustness of LLMs in question-answering tasks.

\subsection{Problem Definition}
Despite the remarkable capabilities of large language models (LLMs), they remain highly sensitive to seemingly trivial input variations—variations that are inconsequential to human understanding. For instance, inserting an additional space after a control keyword such as ``\texttt{Answer:}'' (e.g., ``\texttt{Answer: }'') can lead to substantial drops in exact-match accuracy. These perturbations frequently arise from habitual user phrasing, minor typographical errors, or natural disfluencies in human input. 

Empirical findings reveal that leading LLMs, such as LLaMA and Gemma models, experience notable performance degradation under such conditions. Among them, LLaMA demonstrates relatively stronger robustness, but inconsistencies persist. These results highlight a critical gap in the robustness of current decoding strategies and motivate the need for a decoding-time mechanism capable of mitigating output instability caused by benign input perturbations.

\subsection{Autoregressive Decoding in Large Language Models}
Large language models (LLMs), particularly those based on transformer architectures, generate text in an \textit{autoregressive} manner—predicting one token at a time based on a sequence of previously generated tokens. Formally, given a context \( x_1, x_2, \dots, x_{t-1} \), the model estimates the conditional probability distribution over the next token \( x_t \) as \( P(x_t \mid x_1, x_2, \dots, x_{t-1}) \). The model outputs a distribution over the vocabulary at each step, from which a decoding strategy determines the next token.

One of the most common strategies is \textit{greedy decoding}, where the token with the highest probability is selected at each time step. While efficient and often fluent, greedy decoding can yield repetitive or suboptimal results due to its inability to explore alternative paths. To address this, more diverse decoding methods—such as beam search, top-$k$ sampling, and nucleus (top-$p$) sampling—are frequently used. After each token is generated, it is appended to the context and the process repeats until a stopping criterion is met, such as generating a special end-of-sequence token or reaching a length limit.

\subsection{Token Constraint Decoding (TCD)}
With Token Constraint Decoding, we treat each generation step of an autoregressive language model as an opportunity to enforce hard or soft vocabulary restrictions without altering the model’s parameters. At time step~$t$, the base model produces a logit vector~$\ell^{(t)} \in \mathbb{R}^V$, where $V$ is the vocabulary size. We first convert these logits into a probability distribution via softmax with the equation \ref{equation-distri}:
\begin{equation}
\label{equation-distri}
p_i^{(t)} = \frac{e^{\ell_i^{(t)}}}{\sum_{j=1}^V e^{\ell_j^{(t)}}},
\end{equation}
and then impose a user‑specified constraint set~$A \subseteq \{1, \dots, V\}$ of allowed token indices. Disallowed tokens receive a uniform penalty~$\gamma \ge 0$, which for $\gamma = \infty$ becomes a hard mask (scores set to $-\infty$). We next apply temperature scaling by dividing the penalized probabilities by~$\tau > 0$, yielding adjusted scores that control distribution sharpness.

Rather than sampling at each step, Token Constraint Decoding accumulates these adjusted scores across all~$T$ decoding steps. If $m_i = 1$ for $i \in A$ (else $0$), we compute using the equation \ref{equation-tcd}
\begin{align}
\tilde{p}_i^{(t)} &= p_i^{(t)} - (1 - m_i) \cdot \gamma, \\
q_i^{(t)} &= \frac{1}{\tau} \cdot \tilde{p}_i^{(t)}, \\
S_i^{(t)} &= S_i^{(t-1)} + q_i^{(t)},
\label{equation-tcd}
\end{align}
with initial condition $S_i^{(0)} = 0$. After $T$ steps, the decoder selects the token(s) corresponding to the highest cumulative score $S_i^{(T)}$; for length-$N$ outputs, one takes the top $N$ indices by final score.

This mechanism adds only linear overhead in the vocabulary size per step—one softmax, one mask or subtraction, and one addition for score accumulation—yielding overall complexity~$O(TV)$. Because TCD wraps directly around the model’s logits and introduces no extra parameters or backpropagation paths, it remains lightweight and suitable for real‑time and resource‑constrained settings. By offering a transparent, modular way to ban, restrict, or bias specific tokens, Token Constraint Decoding becomes a versatile tool for controlled, policy‑aware text generation.

\section{Experiment Design}
This work facilitates empirical evaluation of the TCD algorithm's effectiveness. Specifically, it assesses the multiple-choice question-answering capabilities of large language models (LLMs) in commonsense reasoning and language understanding, using established benchmarks such as CommonsenseQA, MMLU, and MMLU Pro. The experiments are structured to isolate the effects of noise, TCD, and prompt engineering (PE) fixes under controlled conditions.

\subsection{Commonsense Reasoning}
To evaluate commonsense reasoning, we use the CommonsenseQA benchmark, which requires models to select the most plausible answer among several distractors based on real-world knowledge. We test four models (Llama3.2 1B, Llama3.2 3B, Llama3.1 7B, and Gemma3 1B Instruct) across six experimental conditions: a clean baseline, noisy input without TCD, and variants with or without TCD and PE fixes. This setup allows us to analyze how reasoning quality is preserved or degraded under noise and whether TCD and prompt-level adjustments can recover model performance. The prompt engineering fix (PE) is to specify the range of output tokens possible in the verbal setting to enable LLMs with better multiple-choice question answering capability.

\subsection{Language Understanding}
Language understanding is assessed using the MMLU and MMLU-Pro benchmarks, which test domain knowledge and linguistic comprehension across academic and professional-level topics. These tasks introduce more abstract and knowledge-heavy queries than CommonsenseQA. We again evaluate all models under the six controlled conditions in our design. The results show that while high-performing models like Llama3.2 3B are somewhat robust to input noise, smaller models such as Llama3.1 7B and Gemma3 1B suffer severe performance degradation without TCD. Again, the PE fixes are facilitated to specify the number of tokens possible for the multiple-choice questions.

\subsection{Relation Between Robustness and Penalty Score}
To further understand how the TCD algorithm improves robustness, we conduct a penalty sweep experiment, varying the penalty score from 0.0 to 1.0 in 0.2 increments. For each penalty level, we evaluate model accuracy on CommonsenseQA, MMLU, and MMLU-Pro to understand its relation to affecting the accuracy score for various benchmarks, including commonsense reasoning and language understanding.

\section{Results}
The result section presents the empirical findings for the above experiments discussed in the previous section. Figures~\ref{fig:commonsenseqa-exp}, \ref{fig:mmlu-exp}, and \ref{fig:mmlupro-exp} present the model accuracies across six experimental conditions: baseline without noise or TCD, baseline with noise, noise with prompt engineering (PE) fix, noise with TCD (w/o PE fix), and full setting with both TCD and prompt engineering (PE) fix. Across all benchmarks, Llama3.2 3B consistently achieves the highest accuracy, confirming its robustness and general capability. In contrast, performance significantly deteriorates across all models under noisy conditions without TCD (especially for Llama3.1 7B and Gemma3), revealing sensitivity to input corruption.

\subsection{CommonsenseQA}
\begin{figure}[ht]
    \centering
    \includegraphics[width=0.8\linewidth]{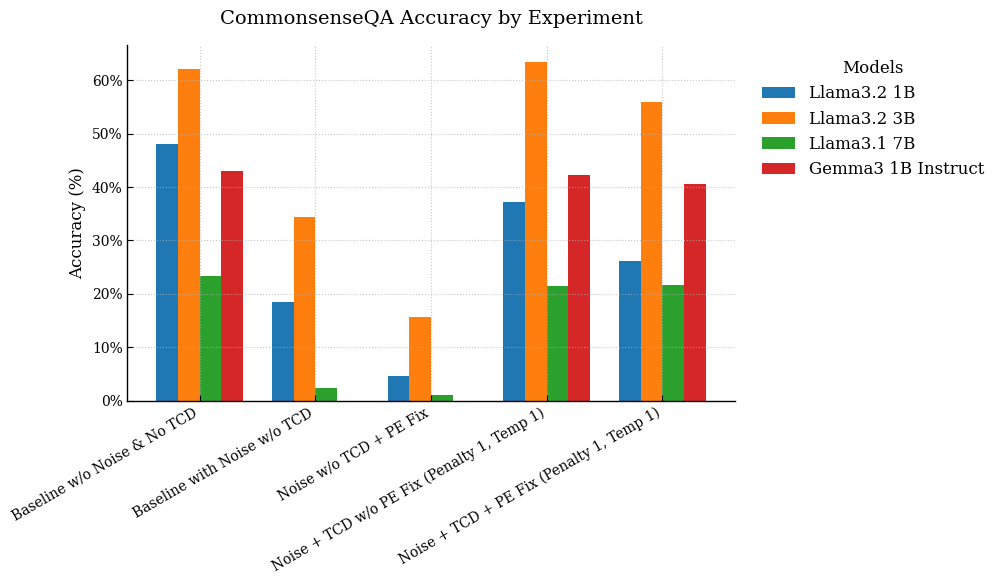}
    \caption{CommonsenseQA accuracy by experiment across four models. TCD + PE Fix recovers most of the performance drop due to noise.}
    \label{fig:commonsenseqa-exp}
\end{figure}

As shown in Figure~\ref{fig:commonsenseqa-exp}, CommonsenseQA accuracy declines steeply when noise is introduced without TCD, especially for smaller models. Llama3.2 3B drops from 62.08\% to 34.31\%, while Gemma3 1B Instruct falls from 42.99\% to 0\%. When only the prompt engineering fix is applied (without TCD), performance remains low, indicating the fix alone is insufficient. Introducing TCD without PE fix improves accuracy, but the best results occur when both TCD and PE fix are used together. In this setting, Llama3.2 3B recovers to 56.09\%, while Gemma3 1B Instruct returns to over 40\%, nearly regaining its original performance. These results suggest that the combination of TCD and PE fix effectively counteracts the impact of noisy inputs.

\subsection{MMLU}
\begin{figure}[ht]
    \centering
    \includegraphics[width=0.8\linewidth]{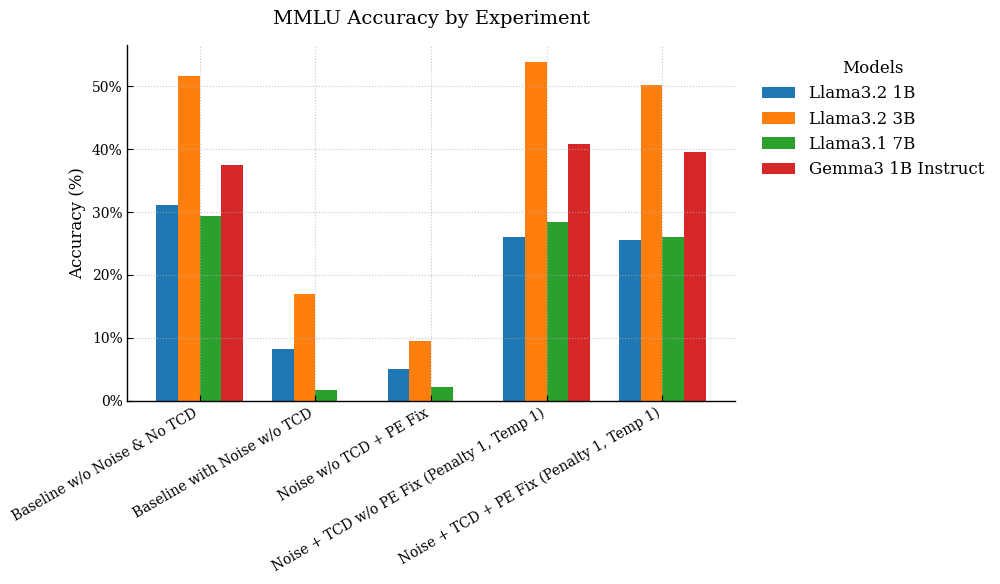}
    \caption{MMLU accuracy by experiment. Llama3.2 3B shows the strongest robustness, while smaller models benefit more from TCD and PE Fix.}
    \label{fig:mmlu-exp}
\end{figure}

Figure~\ref{fig:mmlu-exp} illustrates a similar pattern on the MMLU benchmark. The baseline without noise yields strong performance, with Llama3.2 3B at 51.61\%. However, noise significantly reduces accuracy, especially for smaller or less robust models such as Llama3.1 7B (dropping to 1.66\%) and Gemma3 1B (dropping to 0\%). Once again, the PE fix alone does not suffice to restore performance. The addition of TCD (without PE fix) produces marginal improvements. The best recovery is achieved when both TCD and the PE fix are applied, resulting in substantial accuracy gains across all models. Gemma3 1B Instruct, for example, rises from 0\% (noise only) to 39.02\% in the full setting. This reinforces the importance of using TCD in conjunction with prompt-level calibration.

\subsection{MMLU-Pro}
\begin{figure}[ht]
    \centering
    \includegraphics[width=0.8\linewidth]{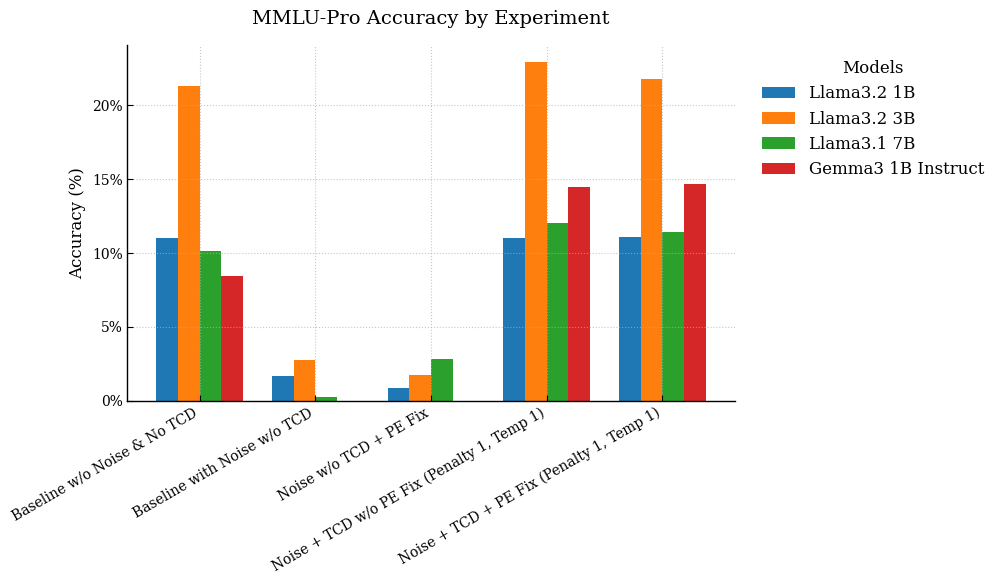}
    \caption{MMLU-Pro accuracy under different conditions. Although more difficult, TCD with PE Fix still improves overall performance.}
    \label{fig:mmlupro-exp}
\end{figure}

As shown in Figure~\ref{fig:mmlupro-exp}, MMLU-Pro represents the most challenging evaluation. All models suffer significant performance degradation under noise, and the overall accuracy values are lower compared to the other benchmarks. Despite this, the combination of TCD and PE fix still provides meaningful gains. Llama3.2 3B, which initially scores 21.31\% without noise, maintains a relatively high 21.87\% accuracy in the full setup, showing its robustness. Gemma3 1B Instruct improves from 0\% to 14.65\% with both fixes applied. While the improvements are less dramatic than on CommonsenseQA or MMLU, the trend remains consistent: TCD and PE fix together are essential for maintaining performance under noisy conditions.

\section{Discussion}
The results presented in this work highlight the practical challenges LLMs face when exposed to input noise and the effectiveness of the Token Consistency Decoding (TCD) algorithm as a mitigation strategy. Despite the growing capabilities of modern LLMs, our findings demonstrate that even state-of-the-art models are vulnerable to performance collapse under perturbed inputs, particularly in structured multiple-choice settings. This vulnerability underscores a critical gap in model robustness and calibration, areas that are often underexplored in favor of raw accuracy on clean datasets.

TCD addresses this gap by introducing a lightweight, inference-time intervention that improves the consistency of model reasoning across token sequences. In combination with prompt engineering fixes, TCD provides a scalable approach to recovering performance without additional training or supervision. In this section, we reflect on key insights from the experiments, analyze where and why TCD is most effective, and discuss its implications for the broader landscape of robust LLM deployment in real-world applications.

\subsection{Effect of Penalty on Model Accuracy}
\begin{figure*}[ht]
    \centering
    \includegraphics[width=\textwidth]{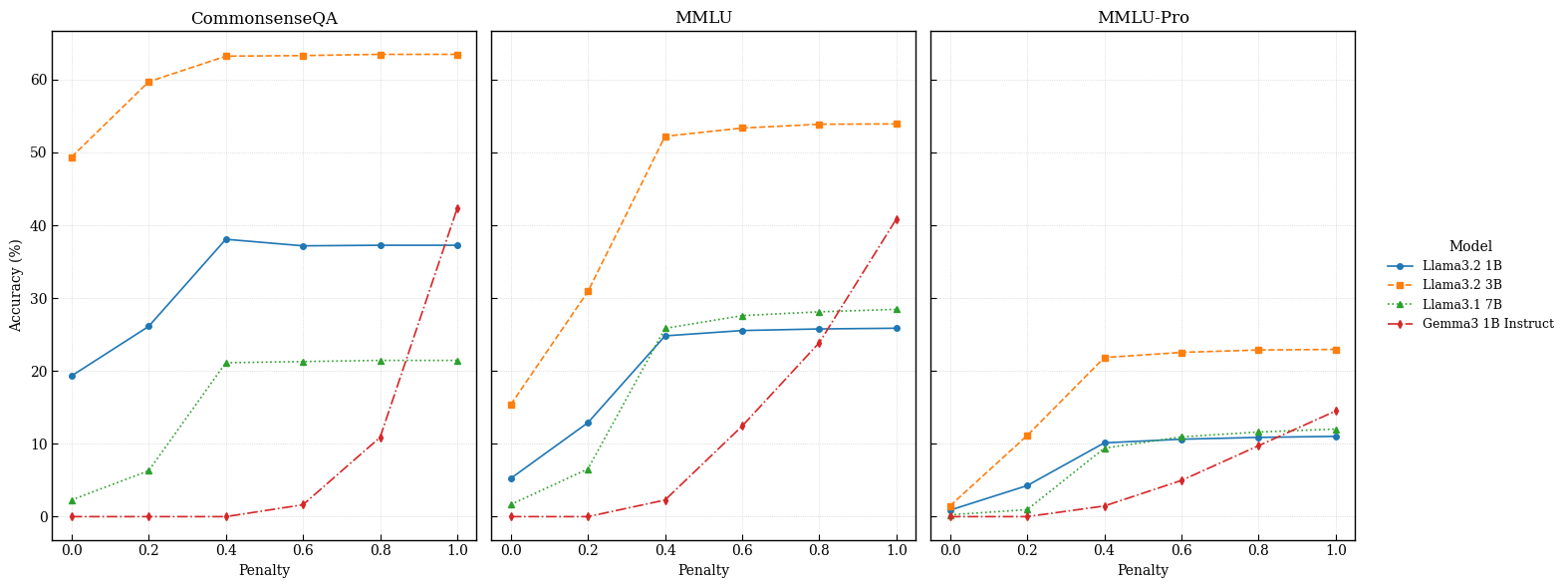}
    \caption{
        Accuracy (\%) of four language models---Llama3.2 1B, Llama3.2 3B, Llama3.1 7B, and Gemma3 1B Instruct---on \textsc{CommonsenseQA}, \textsc{MMLU}, and \textsc{MMLU-Pro} benchmarks across different penalty values (0.0 to 1.0). 
        Llama3.2 3B exhibits consistently strong performance with early saturation, while Gemma3 1B Instruct shows significant improvement only at high penalties.
    }
    \label{fig:penalty-experiment}
\end{figure*}

Figure~\ref{fig:penalty-experiment} illustrates the effect of increasing penalty values on the accuracy of four language models---Llama3.2 1B, Llama3.2 3B, Llama3.1 7B, and Gemma3 1B Instruct---across three benchmark tasks: \textsc{CommonsenseQA}, \textsc{MMLU}, and \textsc{MMLU-Pro}. The x-axis indicates the applied penalty (ranging from 0.0 to 1.0), and the y-axis represents accuracy (\%) on each benchmark. 

Llama3.2 3B consistently achieves the highest performance across all tasks and demonstrates stable improvements with increasing penalties, plateauing at penalty $\geq 0.6$, indicating strong robustness to penalization. Llama3.2 1B shows moderate gains, peaking around penalty 0.4, after which performance saturates or slightly declines, suggesting diminishing returns. Interestingly, Llama3.1 7B---despite having more parameters---underperforms both Llama3.2 models, implying that architectural or training differences may limit its robustness under penalization.

Gemma3 1B Instruct shows the most striking trend: it begins with near-zero accuracy but sharply improves as penalty increases, particularly at penalty = 1.0, where it nearly matches or exceeds Llama3.2 1B on \textsc{CommonsenseQA} and \textsc{MMLU}. This suggests that Gemma3 benefits substantially from regularization, possibly due to overconfidence or high entropy in its initial output distribution.

These results highlight that penalty-based decoding strategies can significantly enhance performance, particularly for weaker or instruction-tuned models. However, the effect is model-dependent, indicating that penalty tuning is a crucial hyperparameter for maximizing performance robustness in multi-choice reasoning tasks.

\subsection{Limitations of the TCD Algorithm}
While Token Constraint Decoding (TCD) offers improved control over LLM outputs in multiple-choice settings, it remains fundamentally constrained by the predefined set of allowable tokens. This design assumes that the correct answer tokens are explicitly enumerable and directly mappable to token-level representations in the model’s vocabulary. Consequently, TCD may struggle in tasks where answer choices are semantically ambiguous, span multiple tokens, or require generative reasoning beyond simple token matching. Moreover, the rigid constraint mechanism can lead to degraded performance if the correct answer is not perfectly aligned with the allowed token set, or if the model is heavily penalized for exploring alternative but valid completions. These limitations suggest that while TCD is effective in structured scenarios, its generalizability to more open-ended or generative tasks remains limited.

\section{Conclusion and Future Work}
This paper presents a comprehensive evaluation of the Token Constraint Decoding (TCD) algorithm for enhancing the robustness of large language models (LLMs) on multiple-choice question answering (MCQA) tasks. Through a series of controlled experiments on CommonsenseQA, MMLU, and MMLU-Pro, we demonstrate that models suffer substantial performance degradation when exposed to input noise. However, the combination of TCD and prompt engineering (PE) fixes significantly restores accuracy, particularly for smaller or instruction-tuned models like Gemma3 1B Instruct. Our results demonstrated that TCD acts as a structural reasoning regularizer, improving alignment and reducing spurious outputs under noisy conditions. Moreover, penalty-sweep experiments reveal that optimal penalty tuning further enhances performance, with underperforming models benefiting most from higher regularization.

\subsection{Future Work}
To address the current limitations of Token Constraint Decoding (TCD), future work will explore extensions that support more flexible and semantically aware decoding strategies. One promising direction is the integration of embedding-based constraints, allowing the model to permit tokens that are semantically similar to valid answers rather than relying solely on exact token matches. This could help mitigate brittleness when tokenization or phrasing differs slightly from the expected form. Another avenue is adaptive constraint relaxation, where the penalty strength \( \lambda \) is dynamically adjusted based on model confidence or contextual cues, allowing for a smoother trade-off between constraint enforcement and generative fluency. Finally, expanding TCD to handle multi-token answer spans and free-form reasoning tasks could extend its applicability beyond multiple-choice formats, potentially through hierarchical decoding or constraint-aware beam search. These directions aim to enhance the robustness and versatility of TCD in real-world LLM deployments.


\bibliographystyle{plainnat}
\bibliography{custom} 


\appendix

\section{Technical Appendices and Supplementary Material}
Technical appendices with additional results, figures, graphs and proofs may be submitted with the paper submission before the full submission deadline (see above), or as a separate PDF in the ZIP file below before the supplementary material deadline. There is no page limit for the technical appendices.


\end{document}